%% file: main.tex
\documentclass[11pt,a4paper]{article}

\usepackage[utf8]{inputenc}
\usepackage[T1]{fontenc}
\usepackage{lmodern}
\usepackage[british]{babel}
\usepackage{microtype}
\usepackage[margin=1in]{geometry}
\usepackage{amsmath,amssymb,amsfonts}
\usepackage{graphicx}
\usepackage{xcolor}
\usepackage{booktabs}
\usepackage{array}
\usepackage{siunitx}
\usepackage{caption}
\usepackage{float}
\usepackage[numbers,sort&compress]{natbib}
\usepackage{enumitem}
\usepackage{tikz}
\usetikzlibrary{arrows.meta,positioning,fit,backgrounds,calc}
\usepackage{url}
\usepackage[colorlinks=true,linkcolor=blue!55!black,citecolor=teal!60!black,%
            urlcolor=blue!55!black,breaklinks=true]{hyperref}
\usepackage{doi}
\hypersetup{
  pdftitle={PLECTA: Overlap-aware instance segmentation of dense strand networks from binary masks},
  pdfauthor={Oday Allan, Igor Ostanin},
  pdfkeywords={strand networks, instance segmentation, graph matching, carbon nanotubes, scanning electron microscopy}
}

\definecolor{fsblue}{HTML}{2B6CB0}
\definecolor{fsorange}{HTML}{DD6B20}
\definecolor{fsgreen}{HTML}{2F855A}

\definecolor{pkink}{HTML}{1A202C}
\definecolor{pksoft}{HTML}{6B7785}
\definecolor{pkacc}{HTML}{2F855A}
\tikzset{
  pk/.style = {
    font=\footnotesize,
    >={Stealth[length=4.5pt,width=3.2pt]},
    every node/.style={inner sep=0pt, outer sep=0pt},
  },
  stage/.style = {
    draw=pkink, line width=.5pt, rounded corners=1.2pt,
    align=center, inner xsep=4pt, inner ysep=5pt,
    text width=#1, minimum height=9.5mm, text=pkink,
  },
  stage/.default = 30mm,
  io/.style = {stage=#1, line width=.9pt},
  io/.default = 30mm,
  ext/.style = {stage=#1, draw=pksoft, text=pksoft},
  ext/.default = 30mm,
  flow/.style = {->, draw=pkink, line width=.5pt},
  flowsoft/.style = {->, draw=pksoft, line width=.5pt},
  loop/.style = {->, draw=pkink, line width=.5pt, dashed,
                 dash pattern=on 2pt off 1.6pt},
  note/.style = {font=\scriptsize, text=pksoft, align=left, inner sep=2pt},
  scope/.style = {
    draw=pkacc, line width=.6pt, dashed, dash pattern=on 2.6pt off 2pt,
    rounded corners=2pt, inner sep=6pt,
  },
  scopelab/.style = {font=\scriptsize, text=pkacc, inner sep=2pt},
}
\newcounter{igorcomment}

\newcommand{\um}{\si{\micro\metre}}
\newcommand{\Fone}{F\textsubscript{1}}

\input{results/plecta_results}



\newcommand{\supprecon}{Supplementary Section~S1}
\newcommand{\suppdepthstage}{Supplementary Section~S2}
\newcommand{\suppmeasures}{Supplementary Section~S3}
\newcommand{\suppcomparators}{Supplementary Section~S4}
\newcommand{\suppdepth}{Supplementary Section~S5}
\newcommand{\suppsplit}{Supplementary Section~S6}
\newcommand{\suppannotation}{Supplementary Section~S7}

\title{\vspace{-1.2cm}\bfseries PLECTA: Overlap-aware instance
segmentation of dense strand networks from binary masks}

\author{%
  \begin{minipage}{0.94\textwidth}
  \centering
  Oday Allan\textsuperscript{a}\quad
  Igor Ostanin\textsuperscript{a,}\thanks{Corresponding author:
  \texttt{i.ostanin@utwente.nl}}\\[5pt]
  {\small\textsuperscript{a}\,Multi-Scale Mechanics (MSM), Faculty of Engineering
  Technology, University of Twente, Enschede, The Netherlands}\\[4pt]
  {\small ORCID: Oday Allan
  \href{https://orcid.org/0000-0002-6341-4695}{0000-0002-6341-4695};\quad
  Igor Ostanin
  \href{https://orcid.org/0000-0003-4233-4255}{0000-0003-4233-4255}}
  \end{minipage}
}
\date{}

\begin{document}
\maketitle

\begin{abstract}
\input{sections/00_abstract}

\end{abstract}


\input{sections/01_introduction}

\input{sections/01b_related_work}
\input{sections/02_materials_methods}
\input{sections/03_results}
\input{sections/04_discussion}

\section*{Funding}
This work was supported by TKI High Tech Systems and Materials (TKI-HTSM) and
the PPS-Innovatieregeling (PPS-I) through the project \emph{Mechanical
Integrity of NanoTube structures} [grant number 24PPS125]. The funders had no
role in study design; in the collection, analysis and interpretation of data;
in the writing of the report; or in the decision to submit the article for
publication.

\section*{Data and code availability}
The PLECTA source, complete fixed configuration, compact seed manifest,
and aggregate machine-readable evidence are available at
\url{https://github.com/Jorallan/PLECTA}; the version used in this paper is
archived at Zenodo \citep{allan2026plectazenodo}. A graphical application that
performs the same segmentation on a supplied mask is available at
\url{https://github.com/Jorallan/PLECTA_APP}. The
annotation tool of Section~\ref{sec:mm-real} is at
\url{https://github.com/Jorallan/ImageAnnotator}; it is listed because the
machine-assisted annotation protocol described there is implemented in it.
The manual annotations of the
\PlectaRealFieldCountLower{} SEM fields, from both annotators and in the form
the scorer reads, will be deposited in 4TU.ResearchData. Generated scenes
and detailed per-scene records are available from the corresponding author
on reasonable request.

\section*{Acknowledgements}
The authors thank Stefan Luding for his helpful comments on the
manuscript.
The authors thank the second annotator (annotator~B of
Section~\ref{sec:mm-real}), who prefers to remain unnamed, for the independent
re-annotation of \PlectaInterAnnotatorFieldCount{} fields on which the
reader-agreement bound in Section~\ref{sec:limitations} rests.
The software used is NumPy, SciPy, scikit-image, NetworkX and Matplotlib, and
the authors thank their maintainers.

\setlength{\bibsep}{0pt plus 0.3ex}
\renewcommand{\bibfont}{\footnotesize}
\bibliographystyle{unsrtnat}
\bibliography{references}

\end{document}

%% file: results/plecta_results.tex
\newcommand{\PlectaHeldoutSceneCount}{128}

\newcommand{\PlectaHeldoutFOneMean}{0.85}
\newcommand{\PlectaHeldoutFOneCILow}{0.84}
\newcommand{\PlectaHeldoutFOneCIHigh}{0.86}

\newcommand{\PlectaDensityTwoZeroFOneMean}{0.91}
\newcommand{\PlectaDensityTwoZeroFOneCILow}{0.89}
\newcommand{\PlectaDensityTwoZeroFOneCIHigh}{0.93}

\newcommand{\PlectaDensitySixZeroFOneMean}{0.77}
\newcommand{\PlectaDensitySixZeroFOneCILow}{0.76}
\newcommand{\PlectaDensitySixZeroFOneCIHigh}{0.79}

\newcommand{\PlectaHeldoutRuntimeMedian}{1.12}

\newcommand{\PlectaHeldoutRuntimeMaximum}{34.06}

\newcommand{\PlectaRobustnessDegradedFOneMean}{0.82}
\newcommand{\PlectaRobustnessCleanFOneMean}{0.84}
\newcommand{\PlectaRobustnessSceneCount}{50}
\newcommand{\PlectaCleanMinusDegradedFOneMean}{0.0224}
\newcommand{\PlectaCleanMinusDegradedFOneCILow}{0.0113}
\newcommand{\PlectaCleanMinusDegradedFOneCIHigh}{0.0333}

\newcommand{\PlectaGreedySceneCount}{50}

\newcommand{\PlectaDNAiCleanSparseFOne}{0.66}

\newcommand{\PlectaRegimeSceneCount}{40}

\newcommand{\PlectaRegimeOverallPlectaFOne}{0.92}

\newcommand{\PlectaRegimeOverallGraFTFOne}{0.48}

\newcommand{\PlectaRegimeOverallDNAiFOne}{0.35}

\newcommand{\PlectaPlacementTolerance}{2}
\newcommand{\PlectaRealFieldCount}{3}
\newcommand{\PlectaRealFieldCountWord}{Three}
\newcommand{\PlectaRealFieldCountLower}{three}
\newcommand{\PlectaBFiveEightOneZeroZeroReferenceCount}{307}

\newcommand{\PlectaBFiveEightOneOneZeroReferenceCount}{430}

\newcommand{\PlectaJoinBFiveEightOneOneZeroUNetPrecision}{0.68}
\newcommand{\PlectaJoinBFiveEightOneOneZeroUNetRecall}{0.53}

\newcommand{\PlectaBFiveEightThreeZeroZeroReferenceCount}{325}

\newcommand{\PlectaRealUNetFOneMean}{0.46}

\newcommand{\PlectaRealManualFOneMean}{0.89}
\newcommand{\PlectaRealSifneFOneMean}{0.87}
\newcommand{\PlectaRealBasuFOneMean}{0.76}

\newcommand{\PlectaInterAnnotatorFieldCount}{3}
\newcommand{\PlectaInterAnnotatorBOnAFOneMean}{0.38}

\newcommand{\PlectaInterAnnotatorAgreeFLow}{0.36}
\newcommand{\PlectaInterAnnotatorAgreeFHigh}{0.53}

\newcommand{\PlectaDepthChance}{0.5}
\newcommand{\PlectaDepthSceneCount}{12}
\newcommand{\PlectaDepthDecidedLow}{0.88}
\newcommand{\PlectaDepthDecidedHigh}{0.96}

\newcommand{\PlectaDepthAllLow}{0.56}
\newcommand{\PlectaDepthAllHigh}{0.81}

\newcommand{\PlectaDepthWithinLow}{0.36}
\newcommand{\PlectaDepthWithinHigh}{0.84}

\newcommand{\PlectaDepthCleanCovTwoZeroLayers}{0.70}

\newcommand{\PlectaDepthCleanCovFourZeroLayers}{0.38}

\newcommand{\PlectaDetectionIoU}{0.1}

%% file: sections/00_abstract.tex
Many materials and tissues, including carbon-nanotube sheets, collagen and
paper, comprise overlapping strands whose arrangement influences their
effective properties. Measuring this geometry from
micrographs requires individual strands to be traced through crossings and
signal gaps, which merge distinct strands or split one strand into disconnected
fragments. We present PLECTA (from Latin \emph{plecta}, a braid), a
deterministic method that segments individual strands using only a binary
image whose foreground pixels mark strand centrelines. At each junction,
PLECTA selects which strand segments continue into one another while
allowing segments to terminate. It formulates
this decision as an exact maximum-weight partial-matching problem, with an
explicit penalty for unpaired segments and no restriction on the number of
segments entering a junction. The output permits crossing pixels to belong to
multiple strands. On synthetic scenes, PLECTA resolved strand identities
more accurately than three released alternatives at every density tested. The
same ranking held on an independently published generator run at its own
settings. A separate stage
uses the corresponding greyscale micrograph to estimate which strand lies on
top at each crossing. The
complete workflow is demonstrated on scanning electron micrographs of
carbon-nanotube sheets, from raw images to individual strand instances with
estimated crossing order, without manual annotations for training or as
pipeline inputs.

%% file: sections/01_introduction.tex
\section{Introduction}
\label{sec:intro}

\begin{figure}[!ht]
  \centering
  \includegraphics[width=\linewidth]{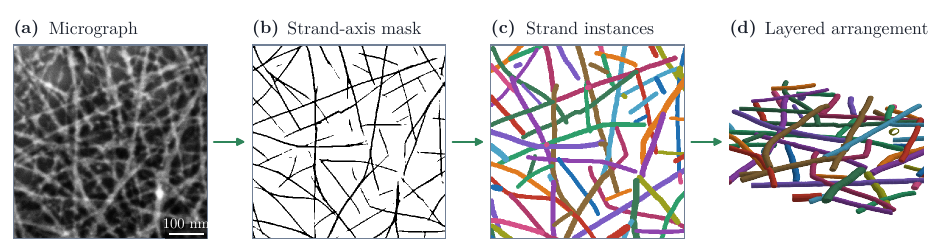}
  \caption{From micrograph to individual strands. \textbf{(a)} Input
    micrograph, a fixed central crop of one carbon-nanotube SEM field.
    \textbf{(b)} Upstream segmenter output. \textbf{(c)} PLECTA
    instance-segmentation output. \textbf{(d)} Depth-stage output.}
  \label{fig:graphical-abstract}
\end{figure}

Carbon-nanotube (CNT) sheets, collagen networks, paper and nonwoven textiles
consist of overlapping strands whose arrangement influences their effective
properties. Their stiffness, strength and conductivity depend on network
geometry, constituent behaviour and inter-strand interactions. Load transfer
between nanotubes, for example, can limit an assembly's stiffness and strength
despite the high modulus of each tube
\citep{devolder2013cnt,coleman2006small,volkov2010structural,%
ostanin2013distinct,ostanin2019size,wittmaack2018mesoscopic}. Predictive models
therefore require
geometric descriptors of individual strands, whose identities must be recovered
from the overlapping network visible in a micrograph.

Recovering those identities is an instance-segmentation problem in which the
objects are long, open curves that overlap one another. Crossings merge
distinct strands into one foreground component, whereas signal gaps split one
strand into fragments whose tips may be indistinguishable from genuine ends.
Segmentation must therefore determine which segments continue through each
junction, which reconnect across a gap and which terminate.

We present a workflow that recovers individual strands and estimates their
depth order from a single greyscale micrograph
(Figure~\ref{fig:graphical-abstract}). First, a segmenter trained on synthetic
image--mask pairs produces a binary mask marking the strand axes, without
requiring manually annotated micrographs for training.

The central segmentation stage, PLECTA, then resolves individual strands from
this mask alone. At each junction, it considers the possible strand
continuations together while allowing genuine strand ends. It also reconnects
fragments across signal gaps. This stage requires no training and is intended
for networks whose strands approximately maintain their direction through
crossings. Finally, a depth-ordering stage uses the segmented strands and
the greyscale image to estimate which strand lies on top at each crossing.
Identity is resolved in the image plane; depth ordering is a separate
downstream stage, and no volumetric reconstruction is performed.

We evaluate PLECTA on synthetic scenes and annotated scanning electron
microscopy (SEM) images of CNT sheets. It achieves higher strand-identity
scores than the compared methods on the evaluated dense-network benchmarks.

Section~\ref{sec:related} reviews related
work, Section~\ref{sec:methods} describes the workflow and evaluation,
Section~\ref{sec:results} presents the results, and
Section~\ref{sec:conclusions} discusses the conclusions and limitations.

%% file: sections/01b_related_work.tex
\section{Related work}
\label{sec:related}

Prior work relevant to the problem of Section~\ref{sec:intro} can be
grouped into five categories: network-level measurement tools, greedy
instance-segmentation methods, joint neighbourhood-optimisation methods,
complete micrograph-to-measurement workflows, and monocular depth-ordering
methods.

\subsection{Network-level measurement tools}
\label{sec:rel-tools}

Existing methods address complementary aspects of the problem. Semantic
segmentation, which labels every pixel with a class but does not distinguish
one object from another, is mature: U-Net, named for its U-shaped
encoder--decoder architecture, and its self-configuring descendants
reliably separate strand foreground from background
\citep{ronneberger2015unet,isensee2021nnunet}, including in materials imaging
\citep{stan2020optimizing,davydzenka2022deep} and nanotube micrographs
specifically \citep{trujillo2017segmentation,nguyen2023cntsegnet,%
dedonato2025deep}. Variants with explicit topology enhancement target the
connectivity of that foreground \citep{lin2026vetunet}, but a foreground mask
still encodes no identity, so it inherits both failures rather than resolving
them. Instance methods go further and
label each object separately, but they assume those objects are compact, in
materials microscopy \citep{he2026instance} as in the general-purpose
families: bounded blobs, centre-directed flows, star-convex outlines,
overlapping ellipses, none of which describes a long open curve that may
overlap another \citep{gu2022instancereview,stringer2021cellpose,%
schmidt2018stardist,kirillov2023sam,panagiotakis2020rfove}. Ridge and
vesselness filters, which score each pixel by how strongly the local image
curvature resembles a line or a tube, sharpen curvilinear signal but assign
nothing across a junction
\citep{steger1998unbiased,frangi1998vesselness,sato1998linefilter}. The
established fibre-analysis packages return network statistics rather than
resolved objects \citep{stein2008fire,bredfeldt2014ctfire,xu2015soax,%
flormann2021finta,hotaling2015diameterj}, as do image-level measures of
coverage and alignment
\citep{dong2022detecting,imtiaz2023facile,beigzadeh2024utilizing}.

Volumetric imaging avoids the problem altogether, because strands that appear
to meet in a projection lie at different depths and are therefore already
apart. X-ray tomography of fibrous solids and of twisted ropes recovers
individual fibres on that basis, by seeded watershed and morphological
reasoning on the reconstructed volume
\citep{depriester2022fibre,allan2025rope}. A single projection offers no such
separation, because two crossing strands genuinely share the same pixels.

\subsection{Greedy continuation from a mask}
\label{sec:rel-greedy}

Recovering strands through crossings has been addressed repeatedly. Existing
methods differ primarily in the evidence used for continuation and in whether
earlier pairings can subsequently be revised.

A large class of methods follows a common procedure: skeletonise the mask,
remove the junction pixels, and pair the resulting arms according to
directional continuity. Each pairing is accepted as soon as it is scored.
SIFNE, an extractor for filament networks in localization-microscopy
images, reads its tip
directions from the excised stubs and recovers dense microtubule networks from
the mask alone \citep{zhang2017sifne}. Wegmayr et al.\ pair skeleton branches
by lowest mutual angle \citep{wegmayr2020microplastic}. GraFT (graph of
filaments over time) covers a Frangi-filtered skeleton greedily, longest path first
\citep{osterlund2026graft}. This is inexpensive, and where crossings are rare
it suffices. Its weakness is structural: once an arm is paired, a later
contradictory candidate cannot revise the pairing.

Two remedies have been tried. The first is to incorporate additional image
evidence beyond the mask. The stronger of Wegmayr et al.'s two variants drops the
angle rule and
learns pixel embeddings on patches around each junction; it reads the greyscale
image, and it must be trained on crops in which every fibre is already labelled
individually. Even then it recovers tangled fibres at about half the precision
of isolated ones, in data where two thirds of fibres cross nothing at all.
Methods that pair ends with an orientation network, detect fibres with a
region-based detector, or trace them recurrently
\citep{liu2019intersection,frei2020fiber,liu2023picktrace} need the same
labelled training data. Reading a mask and nothing else avoids that cost, a
trade-off adopted in this paper. DNAi also uses only the mask but evaluates a larger set of candidate
pairings, enumerating every pairing a junction allows from its endpoint
angles; it does
so exhaustively at degree three and four, and not at all above
\citep{playout2026dnai}. Liu et al.\ also read a mask alone, reconnect gaps and
permit overlapping output, but pair termini greedily and depend on a trained
orientation network \citep{liu2019intersection}.

\subsection{Joint neighbourhood optimisation}
\label{sec:rel-joint}

The second is to resolve several candidate pairings together, rather than
deciding each one alone. Basu et al.\ and
DeFiNe both weigh a whole neighbourhood at once, which is what a dense node
requires. Basu et al.\ delete every branch point from a spanning tree over
centreline points, then reassemble the pieces with an integer program that
penalises each join by its bending, iterating until the assignment
converges; a join that a later fit contradicts can therefore still be undone
\citep{basu2015extraction}. DeFiNe covers an already-built weighted graph with
the smoothest paths it can find, and lets two paths share an edge, so a
crossing pixel can belong to both strands instead of being awarded to one
\citep{breuer2015define}. Basu et al.\ require image intensities, and DeFiNe
requires a graph produced by an upstream stage, rather than an axis mask. Both
also reassemble at most three pieces at a time, which limits how much of a
fragmented neighbourhood a single round can repair.

De et al.\ also weigh a whole neighbourhood at once, but reach a decision by
propagating labels rather than reassembling pieces directly. Branches and
junction
contacts become a graph, and
labels spread across it so that a decision at one node constrains its
neighbours rather than resting on that node's own angles \citep{de2016graph}.
The labels, however, require initialisation. In their images they start at
anatomical roots, the stained cell bodies of neurons or the optic disc of a
retina, and the method returns as many strands as there are roots. A tangle of
open-ended strands offers no such starting point, and how many strands it
holds is precisely the unknown.

\subsection{From micrograph to measurement}
\label{sec:rel-workflows}

Complete workflows separate semantic segmentation from instance segmentation
and validate the combined pipeline against a measurement task. Peters et al.\
join a U-Net-like segmentation trained on over a thousand annotated SEM
micrographs to algorithms that trace fibres through intersections, and
validate against human evaluators on fibre number, length and width
\citep{peters2026fibre}.

Which strand lies on top at a crossing is a further question, and for
nanostructured networks it is normally answered destructively, by
focused-ion-beam sectioning and tomographic reconstruction
\citep{gabbett2024fibsem}, because a micrograph's intensities record surface
topography rather than height. Monocular depth ordering has long derived
pairwise depth from local occlusion evidence, assembled those relations into a
depth-order graph and corrected the contradictions among them
\citep{palou2013monocular,lee2022instancewise}, and single-view hair modelling
resolves the layering of intersecting strands by analogous reasoning
\citep{chai2012singleview}. Amodal instance segmentation also predicts the
hidden extent of each occluded object, guided in DBNet by estimated depth
\citep{zhang2025dbnet}. Those methods read the shape of the outlines where
two objects overlap, which a bundle a few pixels across in a noisy micrograph
does not reliably provide; the evidence used in this paper is instead the
continuity of each strand's own brightness across the crossing.

Weighing a whole neighbourhood at once, with no seed and no strand count
supplied in advance, is the problem addressed in this paper. Four of these
methods are measured against PLECTA directly in
Section~\ref{sec:res-comparators}, on identical masks and through one scorer.

%% file: sections/02_materials_methods.tex
\section{Materials and methods}
\label{sec:methods}

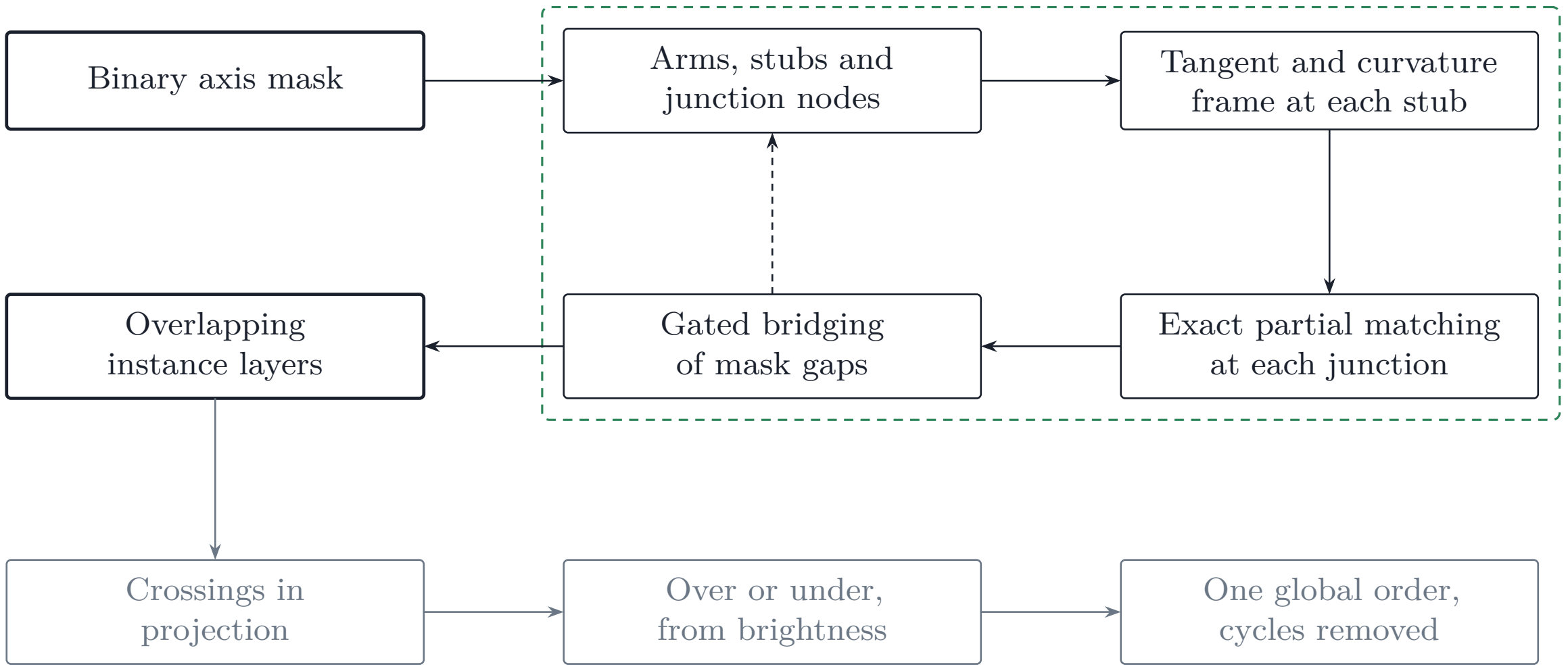
\begin{figure}[!ht]
    \centering
    \input{figures/fig_pipeline_scope}
    \caption{Scope of PLECTA. The dashed grouping stage maps a binary axis mask
    to overlapping instances; the loop denotes eight rounds of frame refitting
    and matching. The grey branch uses a paired greyscale image to infer depth
    order without changing strand identity.}
    \label{fig:scope}
\end{figure}

PLECTA turns a binary strand-axis mask into a set of overlap-aware strand
instances, as shown in the workflow diagram in Figure~\ref{fig:scope}. For
validation and comparison against a known ground truth, the mask can be
generated synthetically; masks derived from SEM images are also tested in this
paper (Sections~\ref{sec:mm-synth} and~\ref{sec:mm-real}). PLECTA then builds a graph
of arms, stubs and junction nodes and fits a tangent and curvature frame to
each stub (Section~\ref{sec:mm-graph}). It scores candidate continuations
(Section~\ref{sec:mm-cost}), solves exact partial matchings at junctions and
across admissible gaps over eight frame-refitting rounds
(Section~\ref{sec:mm-matching}), and converts the accepted links into
overlapping strand instances (Section~\ref{sec:mm-output}). Given a paired
greyscale image, downstream stages estimate strand width and render ribbons
(Section~\ref{sec:mm-stage4}) and infer depth order at crossings
(Section~\ref{sec:mm-depth}). Section~\ref{sec:mm-metrics} defines the
measures all results are scored through and their statistical treatment.

\subsection{Synthetic data}
\label{sec:mm-synth}

Strand centrelines are generated as random walks with Gaussian-smoothed
angular increments, producing gradual bends with controllable curvature.
The principal synthetic segmentation scenes are $512\times512$ pixels.
Each full-width bundle is assigned a width sampled uniformly from 7 to 16
pixels, and strands are placed at random positions and orientations until
their combined full-width area reaches the target areal coverage of 20, 30,
40, 50 or 60\,\% of the image area. Each scene provides strand centrelines
$\mathcal C_k$, their corresponding pixel sets $\Omega_k$, and four image
representations: a clean one-pixel axis mask; a one-pixel axis mask degraded
with gaps and raster irregularities; a full-width foreground mask; and an
SEM-like greyscale image with
strand-specific brightness, gradual intensity variation, blur and background
noise. For depth evaluation (Section~\ref{sec:mm-depth}), a depth-resolved
variant assigns strand heights and radii, prevents interpenetration and
renders the scene from above. Occlusion, depth-dependent defocus and width
broadening are included, with imaging cues randomised independently so that
no single cue reliably determines crossing order.

\subsection{CNT SEM data and mask sources}
\label{sec:mm-real}

\PlectaRealFieldCountWord{} carbon-nanotube (CNT) sheet fields, imaged by
scanning electron microscopy (SEM), were analysed, each comprising
$1536\times1024$ pixels: B58\_100, B58\_110 and B58\_300, containing
\PlectaBFiveEightOneZeroZeroReferenceCount{},
\PlectaBFiveEightOneOneZeroReferenceCount{} and
\PlectaBFiveEightThreeZeroZeroReferenceCount{} annotated bundles, respectively.
Each strand annotation represents a visible bundle, as individual nanotubes
are unresolved at this magnification. The horizontal field width was
$1.66~\um$, corresponding to $1.08~\mathrm{nm\,px^{-1}}$. Images were acquired
using a Thermo Fisher Helios 5~UX DualBeam with an Elstar field-emission column
in immersion mode and a through-lens detector, at 2.00\,kV, 400\,pA and a
working distance of 4.3\,mm. Each frame integrated 35 passes at a dwell time
of 100\,ns with drift correction.

The first author (O.A.) and an independent second annotator each annotated
all \PlectaInterAnnotatorFieldCount{} fields using an in-house tool that permits
overlapping strand masks. Ridge snapping and spline smoothing assisted
tracing, while strand connections and suggested gap bridges required manual
decisions. Scores are reported separately against each annotator, with their
agreement discussed in
Section~\ref{sec:limitations}. \suppannotation{} describes the
annotation procedure.

Two input masks were evaluated for each field: a skeletonised mask derived
from the first author's reference and an automatically generated mask from the
upstream CycleGAN (cycle-consistent generative adversarial network) and
nnU-Net segmentation (Section~\ref{sec:res-qualitative})
\citep{ruehle2021workflow,isensee2021nnunet}. These inputs assess segmentation
from manually curated and automatically extracted strand axes, respectively.
The automatic segmentation required no manual annotations for training or
inference.

Strand segmentation was evaluated on the common centreline fragments
visible within each input mask,
using the metrics defined in Section~\ref{sec:mm-metrics}.

\subsection{PLECTA instance segmentation}
\label{sec:mm-plecta}

Starting from a binary strand mask, PLECTA segments individual strands by
resolving connections through junctions and across gaps. The following
subsections describe the main steps: constructing the skeleton graph, scoring
candidate connections and selecting compatible links, with iterative
refinement as longer strands are assembled. For an input mask
$M\in\{0,1\}^{H\times W}$, the output is a set of strand instances
$\{I_1,\ldots,I_N\}$, with $I_n\subseteq F=\{x:M(x)=1\}$; crossing pixels
may belong to multiple instances. Segmentation uses the mask alone, while
subsequent width and depth estimates also use the greyscale image
(Section~\ref{sec:mm-image}).

\subsubsection{Skeleton graph and local geometry}
\label{sec:mm-graph}

PLECTA skeletonises the input mask using the Zhang--Suen thinning algorithm
\citep{zhang1984thinning} and removes short spurs. Pixels where three or more
skeleton branches meet are grouped into junction clusters, with nearby
clusters merged into a single \emph{node}. This accounts for crossings that
span several pixels or lie too close together to resolve separately. The
remaining strand segments form \emph{arms}, each with a \emph{stub} at either
end.

Each stub is characterised by its tip position, outward tangent
($\mathbf{t}$) and signed curvature ($\kappa$), estimated from a short segment
of the adjoining strand (Figure~\ref{fig:cost}a). Subscripts identify the stub,
so $\mathbf{t}_a$ and $\kappa_a$ denote the tangent and curvature of stub $a$.
Stubs with insufficient support for a reliable direction estimate are marked
as unreliable. As arms are linked into longer chains, these estimates are
updated using the additional strand geometry.

\subsubsection{Pairing cost function}
\label{sec:mm-cost}

\begin{figure}[tbp]
    \centering
    \includegraphics[width=\linewidth]{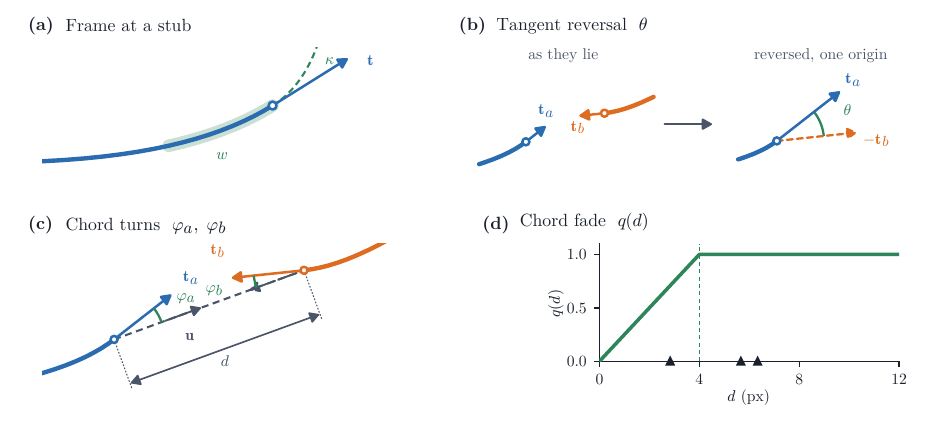}
    \caption{The stub frame and the terms of the pairing cost.
    \textbf{(a)}~The least-squares arclength fit over window $w$, read outward
    from the tip at $s=0$, giving outward tangent $\mathbf{t}$ and signed
    curvature $\kappa$; quadratic when the span allows, linear otherwise.
    \textbf{(b)}~$\theta$, in the two steps its definition takes: the outward
    tangents where they lie, then the same pair with $\mathbf{t}_b$ reversed and
    both read from one origin. A perfect continuation makes them antiparallel,
    so $\theta=0$. \textbf{(c)}~The chord turns $\varphi_a$, $\varphi_b$ onto the
    chord $\mathbf{u}$ joining the tips. \textbf{(d)}~The fade $q(d)$ against tip
    separation; triangles mark the six real separations at the crossing of
    Figure~\ref{fig:matching}.}
    \label{fig:cost}
\end{figure}

The cost $C_{ab}$ of linking stubs $a$ and $b$ combines directional alignment,
chord alignment and curvature continuity (Equation~\ref{eq:cost};
Figure~\ref{fig:cost}). The reversal angle $\theta$, measured between
$\mathbf{t}_a$ and $-\mathbf{t}_b$, is zero when the outward tangents form a
straight continuation. The chord angles $\varphi_a$ and $\varphi_b$ measure
how far each tangent turns towards the opposite tip, penalising lateral
offset between otherwise aligned arms. The curvature mismatch
$\lvert\kappa_a+\kappa_b\rvert$ penalises changes in bending; the sum reflects
the opposite orientations of the outward frames. Gap links also incur a
penalty for tip separation $d$:
\begin{equation}
  C_{ab}=w_d\,\theta
        +w_t\,q(d)\,(\varphi_a+\varphi_b)
        +w_\ell\,\frac{d}{\ell_0}
        +30\,w_\kappa\,\lvert\kappa_a+\kappa_b\rvert,
  \qquad q(d)=\min\!\left(1,\tfrac{d}{d_0}\right),
  \label{eq:cost}
\end{equation}
where $w_d$, $w_t$, $w_\ell$ and $w_\kappa$ weight the four terms, and
$\ell_0$ normalises the gap distance. Junction links set $w_\ell=0$.

The reversal angle $\theta$ is independent of the tip-to-tip chord and remains
useful when nearby tips make the chord direction sensitive to pixel
quantisation. The factor $q(d)$ therefore reduces the chord-angle contribution
at separations below $d_0$. The weights, distance scales $\ell_0$ and $d_0$,
and curvature scaling factor of 30 are fixed parameters listed in
\supprecon{}. These costs quantify the suitability of individual connections;
the next step selects a compatible set of links.

\subsubsection{Candidate selection and iterative matching}
\label{sec:mm-matching}

Using the costs from Equation~\ref{eq:cost}, PLECTA selects links jointly so
that no stub is assigned to more than one connection. Leaving stub $i$
unmatched incurs a penalty $p_i$. Connecting stubs $i$ and $j$ avoids both
unmatched penalties but incurs the linking cost $C_{ij}$ defined in
Equation~\ref{eq:cost}. The net benefit of the connection is therefore
$p_i+p_j-C_{ij}$. A link is admissible only if $C_{ij}<p_i+p_j$.
Exact maximum-weight partial matching then selects the set of disjoint stub
pairs $\Gamma$ with the greatest total benefit:
\begin{equation}
  \max_{\Gamma}\sum_{(i,j)\in\Gamma}
  \bigl(p_i+p_j-C_{ij}\bigr).
  \label{eq:matching}
\end{equation}
Each stub can belong to at most one pair or remain unmatched, allowing genuine
strand termination at junctions of any degree. The penalties therefore
determine both candidate admissibility and the preference for linking. Because
competing links are considered together, a locally favourable pair may be
rejected when another combination gives a better overall assignment.

\begin{figure}[tbp]
    \centering
    \includegraphics[width=\linewidth]{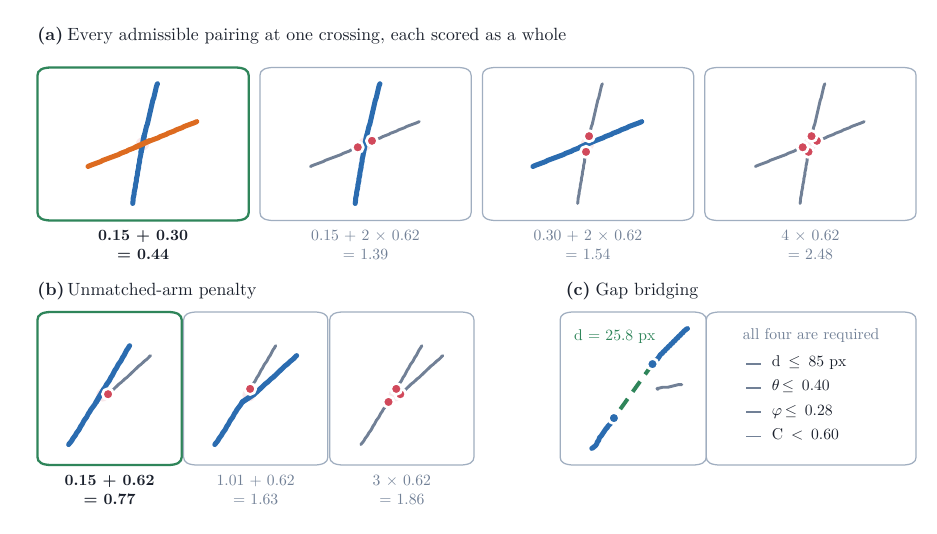}
    \caption{Candidate-link evaluation and matching outcomes, on the nodes
    where the decisions were made. Every arm is the skeleton the numbers were
    measured from, so every angle shown is the one the cost was computed on.
    \textbf{(a)}~Every pairing the admissibility gate allows at a four-arm
    crossing, each scored as a whole: its edge costs plus the penalty
    $p_i=0.62$ of each arm it leaves unmatched. The green frame marks the one
    returned; two further pairs exceed $p_i+p_j=1.24$ and are excluded as
    candidates. \textbf{(b)}~A three-arm node where an admissible edge is
    declined on the total rather than on its own cost. \textbf{(c)}~A gap link,
    which must clear all four gates at once.}
    \label{fig:matching}
\end{figure}

Junction and gap links use different penalties. Gap candidates must also
satisfy distance and angular limits, since they connect strands across missing
image evidence. Figure~\ref{fig:matching} illustrates candidate selection and
matching at junctions and across gaps.

Matching proceeds iteratively, with tangent and curvature estimates updated
along the assembled chains before each round. Early rounds use lower unmatched
penalties and tighter gap-angle limits to restrict uncertain connections.
These constraints gradually relax to their configured values, and the final
round determines the output. \supprecon{} lists the penalties, geometric
limits and iteration schedule; Section~\ref{sec:res-ablation} evaluates joint
matching and the contributions of individual components.

\subsubsection{Strand instance assembly}
\label{sec:mm-output}

After the final matching round, connected components of the accepted arm
links define individual strands. Each strand contains the pixels of its
constituent arms and associated junction clusters, allowing intersecting
strands to share crossing pixels. Accepted gap links connect fragments
belonging to the same strand without adding pixels across the gap. These
strand instances provide the input to the image-based stages described next.

\subsection{Image-informed stages}
\label{sec:mm-image}

The segmented strands are combined with the corresponding greyscale image
to estimate their apparent widths and render ribbons
(Section~\ref{sec:mm-stage4}), and to infer crossing order
(Section~\ref{sec:mm-depth}).
These stages preserve the strand identities established during segmentation
and do not affect the identity scores. They use fixed parameters
(\supprecon{}) and are demonstrated here on SEM images, although the procedure
accepts other greyscale images.

\subsubsection{Width and ribbon rendering}
\label{sec:mm-stage4}

Intensity profiles sampled perpendicular to each strand, away from junctions,
provide estimates of apparent width and brightness, together with uncertainty
and quality indicators. The centreline is then resampled and smoothed, with
accepted gaps interpolated, to render a ribbon at the estimated width.
Junction clusters are excluded from this rendering, and strands remain
separate. The resulting ribbons describe their projected shapes; their
relative positions in depth are estimated in Section~\ref{sec:mm-depth}.

\subsubsection{Depth ordering and height assignment}
\label{sec:mm-depth}

Depth estimation proceeds in three steps: estimating local crossing orders,
combining them into a consistent strand order, and assigning layers and heights.

\begin{figure}[!ht]
    \centering
    \includegraphics[width=\linewidth]{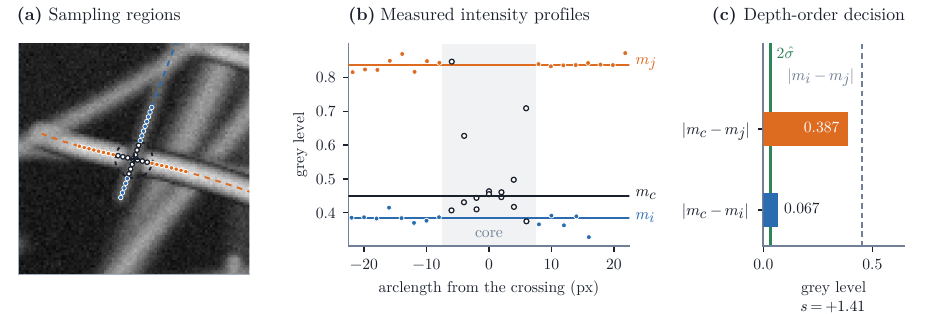}
    \caption{Local depth evidence at a crossing angle of $83.8^{\circ}$,
    selected for clarity. Positions and intensities are taken from the stage
    output. \textbf{(a)}~Core samples (open dots) and flank samples (filled
    dots), with flanks kept clear of the other strand.
    \textbf{(b)}~Intensity along each strand: the pooled core median $m_c$
    is closer to strand $i$'s flank median than to strand $j$'s, supporting
    $i$ as the upper strand. \textbf{(c)}~Intensity mismatches used to
    calculate $s$, normalised by the larger of the flank-median separation
    and twice the local noise estimate (\suppdepthstage{}).}
    \label{fig:depth-evidence}
\end{figure}

\textbf{1. Estimate crossing order.} The method assumes that the upper strand
retains its brightness through a crossing, while the lower strand is occluded.
Intensity is sampled through the crossing (the \emph{core}) and along each
strand on either side (the \emph{flanks}), excluding flank pixels occupied by
the other strand (Figure~\ref{fig:depth-evidence}a,b). The pooled core median
$m_c$ is compared with the flank medians $m_i$ and $m_j$; the closer match
indicates the upper strand.

The signed score $s$ summarises this comparison
(Figure~\ref{fig:depth-evidence}c). It compares the two intensity mismatches,
normalised by the larger of $\lvert m_i-m_j\rvert$ and twice the local noise
estimate $\hat\sigma$. Its sign identifies the upper strand and its magnitude
weights the evidence. Crossings with insufficient flank samples or a score
below the threshold in magnitude remain undecided. \suppdepthstage{} gives
the sampling procedure and score equation.

\begin{figure}[!ht]
    \centering
    \includegraphics[width=\linewidth]{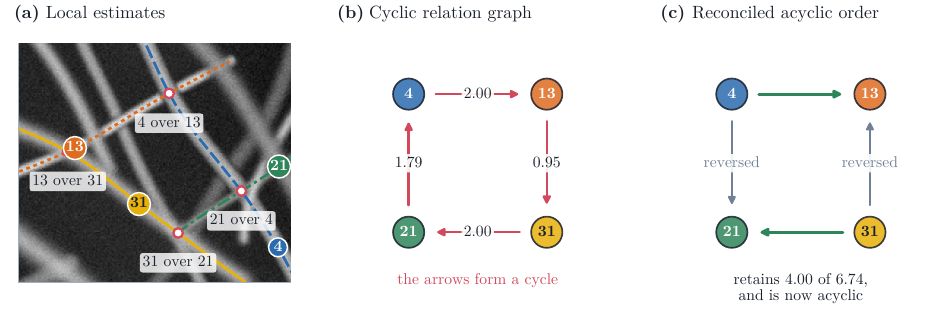}
    \caption{Resolving contradictory local depth estimates.
    \textbf{(a)}~Four strands and their estimated crossing orders.
    \textbf{(b)}~The corresponding graph, with arrows from the estimated
    upper strand to the lower strand. The directed cycle prevents a
    consistent height assignment. \textbf{(c)}~The selected order retains
    the two highest-weight estimates and reverses the two lowest-weight
    estimates. Reversals are recorded with their original scores. The four
    strands shown belong to a seven-strand scene; layer assignments depend
    on the complete connected component.}
    \label{fig:depth-order}
\end{figure}

\textbf{2. Combine crossing orders.} Local estimates can conflict: $A$ may
lie above $B$, $B$ above $C$, and $C$ above $A$. These relations form a directed
graph whose nodes are strands and whose edges carry weights $\lvert s\rvert$
(Figure~\ref{fig:depth-order}). Within each connected component, a consistent
strand order is sought with the objective of maximising the total weight of
satisfied relations. Crossing orders are updated accordingly.
\suppdepthstage{} describes the ordering procedure.

\textbf{3. Assign layers and heights.} The corrected relations form a graph
without directed cycles. Its longest path determines the minimum number of
layers needed to respect the crossing orders; strands without ordering
constraints between them may share a layer. Under the compact-stack
assumption, each strand is assigned the lowest height compatible with the
chosen order and the estimated radii, preventing interpenetration.

Without a greyscale image, placement follows the predefined tie-break rather
than measured crossing evidence. Section~\ref{sec:res-depth} evaluates depth ordering
using the measures defined in \suppdepth{}.

\subsection{Metrics and statistical analysis}
\label{sec:mm-metrics}

\begin{figure}[!ht]
    \centering
    \includegraphics[width=\linewidth]{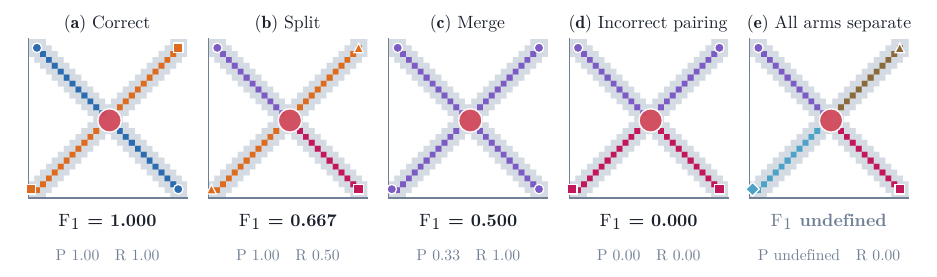}
    \caption{Pairwise scoring at a crossing. Removing junction pixels (red)
    leaves four arms and six fragment pairs. Tip markers identify arms assigned
    to the same strand. \textbf{(a)}~Reference grouping.
    \textbf{(b)}~A split reduces recall. \textbf{(c)}~A merge reduces precision.
    \textbf{(d)}~Incorrect pairing changes strand identities.
    \textbf{(e)}~Separating every arm produces no predicted positive pair, so
    precision and \Fone{} are undefined.}
    \label{fig:f1-example}
\end{figure}

The primary segmentation metric is common-fragment pairwise \Fone{}, which
measures whether strand fragments are grouped consistently with the reference.
Each input mask is divided into fragments between crossings, independently of
the reference and predictions. The same assignment rule then gives each
fragment a reference identity and a predicted identity. Pairwise precision
penalises joining fragments from different reference strands, while recall
penalises separating fragments from the same strand.
Figure~\ref{fig:f1-example} illustrates these cases and the conditions under
which the score is undefined.

Because junction pixels are excluded, this metric evaluates strand continuity
through crossings rather than ownership of the crossing pixels themselves.
Strands containing more fragments contribute more pairs and therefore carry
greater weight. We also report the adjusted Rand index and variation of
information \citep{hubert1985comparing,meila2007comparing}, together with
detection \Fone{} and Crossing Fidelity, to characterise complementary aspects
of segmentation. For the real SEM fields, Join \Fone{} evaluates joining
decisions between nearby fragments, reducing the influence of differences in
fragmentation between mask sources.

Synthetic segmentation means are accompanied by 95\,\% bootstrap
confidence intervals, with scenes resampled within each density level.
Comparisons between clean and degraded masks preserve scene pairing. For
real SEM data, each image field is treated as one independent sample.
\suppmeasures{} provides the metric definitions, assignment rules and
statistical procedures.

%% file: figures/fig_pipeline_scope.tex
\begin{tikzpicture}[pk, x=1mm, y=1mm]

  \node[io=38mm]    (mask)   at ( 27,   0) {Binary axis mask};
  \node[stage=38mm] (graph)  at ( 81.5, 0) {Arms, stubs and\\junction nodes};
  \node[stage=38mm] (frame)  at (136,   0) {Tangent and curvature\\frame at each stub};

  \node[io=38mm]    (out)    at ( 27,  -26) {Overlapping\\instance layers};
  \node[stage=38mm] (bridge) at ( 81.5,-26) {Gated bridging\\of mask gaps};
  \node[stage=38mm] (match)  at (136,  -26) {Exact partial matching\\at each junction};

  \draw[flow] (mask)   -- (graph);
  \draw[flow] (graph)  -- (frame);
  \draw[flow] (frame)  -- (match);
  \draw[flow] (match)  -- (bridge);
  \draw[flow] (bridge) -- (out);

  \draw[loop] (bridge.north) -- (graph.south);

  \begin{scope}[on background layer]
    \node[scope, fit=(graph)(frame)(match)(bridge)] (core) {};
  \end{scope}

  \node[ext=38mm] (cross) at ( 27,  -52) {Crossings in\\projection};
  \node[ext=38mm] (ou)    at ( 81.5,-52) {Over or under,\\from brightness};
  \node[ext=38mm] (order) at (136,  -52) {One global order,\\cycles removed};

  \draw[flowsoft] (cross) -- (ou);
  \draw[flowsoft] (ou)    -- (order);
  \draw[flowsoft] (out.south) -- (cross.north);

\end{tikzpicture}

%% file: sections/03_results.tex
\section{Results}
\label{sec:results}

We first evaluate strand segmentation on synthetic scenes
(Section~\ref{sec:res-density}), then examine depth ordering
(Section~\ref{sec:res-depth}) and application to real SEM images
(Section~\ref{sec:res-real}). Component ablations
(Section~\ref{sec:res-ablation}) and comparisons with published methods
(Section~\ref{sec:res-comparators}) assess the contributions and performance
of the segmentation method.

\subsection{Instance segmentation on synthetic images}
\label{sec:res-density}

These scenes carry exact reference identities, so every junction and gap
decision can be checked against the grouping the generator recorded when it
drew them.
Across the \PlectaHeldoutSceneCount{} synthetic scenes, PLECTA
achieved a mean common-fragment pairwise \Fone{} of
\PlectaHeldoutFOneMean{} (95\% bootstrap CI
[\PlectaHeldoutFOneCILow{}, \PlectaHeldoutFOneCIHigh{}]).
Table~\ref{tab:plecta-heldout} summarises segmentation scores overall and by
areal coverage.

\begin{table}[tbp]
    \centering
    \caption{Segmentation scores overall and by target areal
    coverage. VI denotes variation of information, for which lower is better;
    higher is better for the other metrics.}
    \label{tab:plecta-heldout}
    \small
    \input{results/plecta_heldout_table}
\end{table}

Performance decreased with scene density: mean \Fone{} fell from
\PlectaDensityTwoZeroFOneMean{}
[\PlectaDensityTwoZeroFOneCILow{}, \PlectaDensityTwoZeroFOneCIHigh{}] at 20\%
coverage to \PlectaDensitySixZeroFOneMean{}
[\PlectaDensitySixZeroFOneCILow{}, \PlectaDensitySixZeroFOneCIHigh{}] at 60\%,
monotonically through the intermediate strata.
Per-scene runtime had median \PlectaHeldoutRuntimeMedian{}~s, ranging to
\PlectaHeldoutRuntimeMaximum{}~s on the densest scenes;
Section~\ref{sec:res-comparators} measures cost systematically.

PLECTA operates on the strand-axis mask, so segmentation difficulty depends
on centreline density and geometry rather than areal coverage alone. The same
areal coverage can arise from a different centreline density under another
bundle-width distribution and may therefore produce different segmentation
performance. Within the fixed $512\times512$-pixel frame and 7--16-pixel
bundle-width range used here, areal coverage provides a consistent within-study
stratification.

\label{sec:res-mask}%
To assess sensitivity to gaps and raster irregularities, we compared
segmentation from clean and degraded masks of the same scenes. Across the
\PlectaRobustnessSceneCount{} synthetic scenes,
mean \Fone{} was \PlectaRobustnessDegradedFOneMean{} from degraded axes and
\PlectaRobustnessCleanFOneMean{} from the corresponding clean axes. The paired
clean-minus-degraded change was \PlectaCleanMinusDegradedFOneMean{} (95\% CI
[\PlectaCleanMinusDegradedFOneCILow{},
\PlectaCleanMinusDegradedFOneCIHigh{}]). Modelled gaps and raster
irregularities therefore caused a small reduction in segmentation accuracy
under this generator. The difference was greatest at low coverage and narrowed
at 50--60\,\%, consistent with junction ambiguity becoming more influential
at higher density.

\begin{figure}[tbp]
    \centering
    \includegraphics[width=\linewidth]{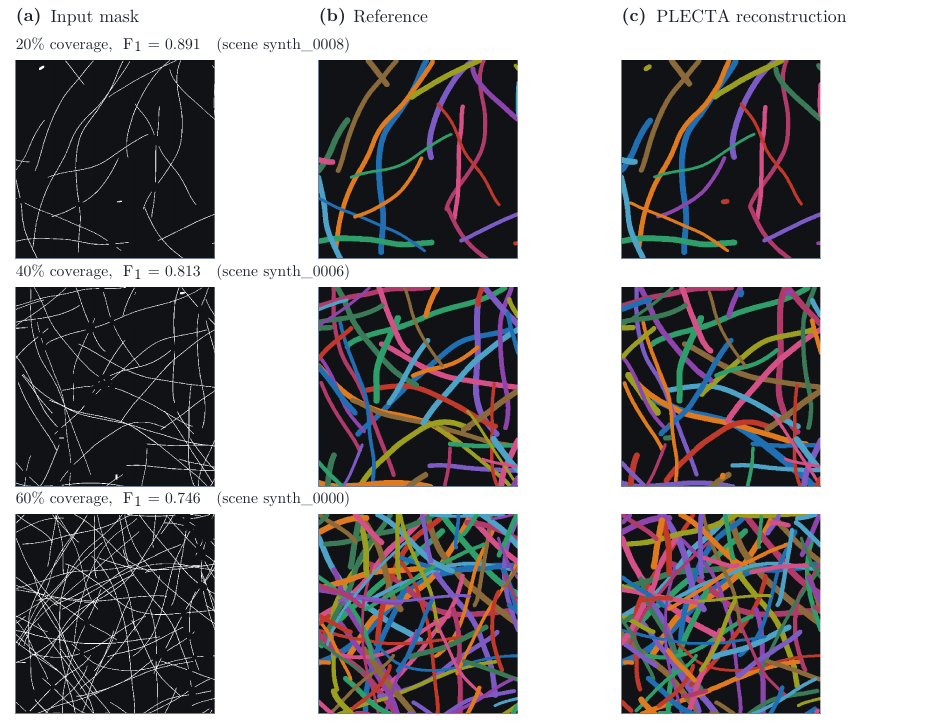}
    \caption{Segmentation at three coverage levels, each the scene closest
    to the median \Fone{} of its stratum, shown as a horizontal window of the
    scene. Columns: the degraded input mask, the overlap-aware reference, and
    PLECTA's strands at their fitted widths. The score beside each row is the
    grouping's, not the rendering's. Strand colours are arbitrary and not
    matched between columns.}
    \label{fig:examples}
\end{figure}

Figure~\ref{fig:examples} shows representative segmentations at three
coverage levels, selected by proximity to the median \Fone{} of each stratum.

\subsection{Depth decomposition}
\label{sec:res-depth}

The stages above segment strands in projection but do not establish which
strand lies above at a crossing. After grouping is fixed, the depth stage uses
the greyscale image to assess the continuity of each strand's brightness across
the crossing relative to local noise. It reconciles these local estimates into
a globally consistent vertical order and assigns non-interpenetrating heights.

The stage was evaluated on \PlectaDepthSceneCount{} depth-resolved scenes from
Section~\ref{sec:mm-synth}, at two areal coverages and under two input
conditions. Reference 2-D geometry isolates the depth decision, whereas binary
axis masks include segmentation error. Table~\ref{tab:depth-order} reports
both using the measures defined in \suppdepth{}. Its crossing rows score local
over/under decisions before reconciliation, once for every crossing; its
stacking rows score the final order, once for every pair of strands.

\begin{table}[!ht]
\centering
\caption{The depth stage on \PlectaDepthSceneCount{} synthetic scenes whose
true vertical order is known. Chance is \PlectaDepthChance{} for every
accuracy shown.}
\label{tab:depth-order}
\input{results/plecta_depth_order_table}
\end{table}

Among crossings for which an order was assigned, accuracy ranged from
\PlectaDepthDecidedLow{} to \PlectaDepthDecidedHigh{} across both input
conditions and densities. The all-crossing scores were lower
(\PlectaDepthAllLow{}--\PlectaDepthAllHigh{}) because they include crossings
the 2-D stage did not recover and those for which the depth stage abstained.
The table reports recovery and abstention rates alongside these scores.

Global order is limited to evidence that can travel through the crossing graph.
Only \PlectaDepthWithinLow{} to \PlectaDepthWithinHigh{} of strand pairs lie
within the same connected component, so the relative order of the remaining
pairs is not identifiable from the available crossing evidence. All-pair order
accuracy was consequently 0.67--0.72, compared with 0.85--0.91 for pairs that
cross. Layer agreement likewise fell from \PlectaDepthCleanCovTwoZeroLayers{}
to \PlectaDepthCleanCovFourZeroLayers{} as coverage doubled, although the
recovered number of layers remained close to the reference.

\begin{figure}[!ht]
\centering
\includegraphics[width=\linewidth]{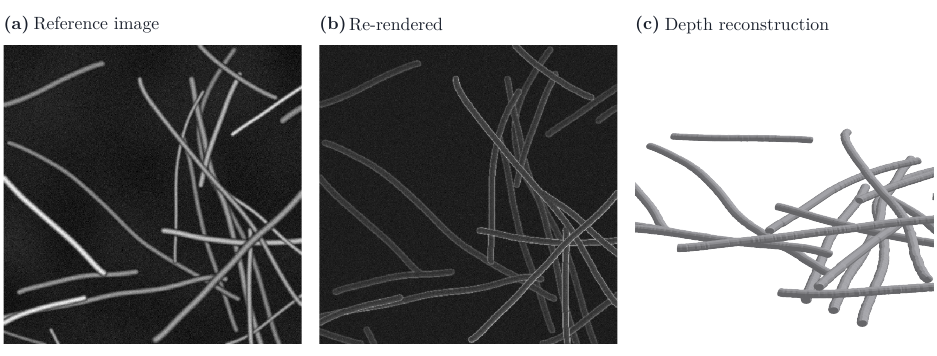}
\caption{Depth reconstruction of a synthetic scene from its binary axis mask.
The panels show the rendered greyscale image, a forward-model re-rendering from
the reconstructed depths and radii, and the inferred strand stack. The
re-rendering is a visual consistency check, not a scored quantity. The scene is
512\,px across in the generator's pixel units.}
\label{fig:resem-synth}
\end{figure}

\begin{figure}[tbp]
\centering
\includegraphics[width=\linewidth]{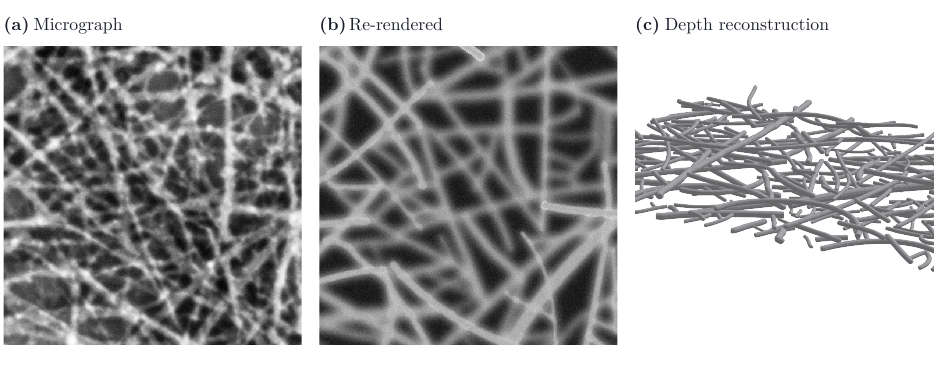}
\caption{The same comparison on a real CNT SEM field, reconstructed from the
reference axis mask. The crop is 512\,px across, about 0.55\,\um{} of field
B58\_100. Real SEM provides no depth ground truth, so the depth order is shown
qualitatively; the forward-model appearance parameters are fitted to this
micrograph.}
\label{fig:resem-real}
\end{figure}

Figure~\ref{fig:resem-synth} shows what the stage produces on one synthetic
scene, from the binary axis mask through to the film the reconstruction
implies; Figure~\ref{fig:resem-real} is the same sequence on a real CNT SEM
field, which no depth ground truth can score.

\subsection{Instance segmentation on CNT SEM images}
\label{sec:res-real}
\label{sec:res-qualitative}

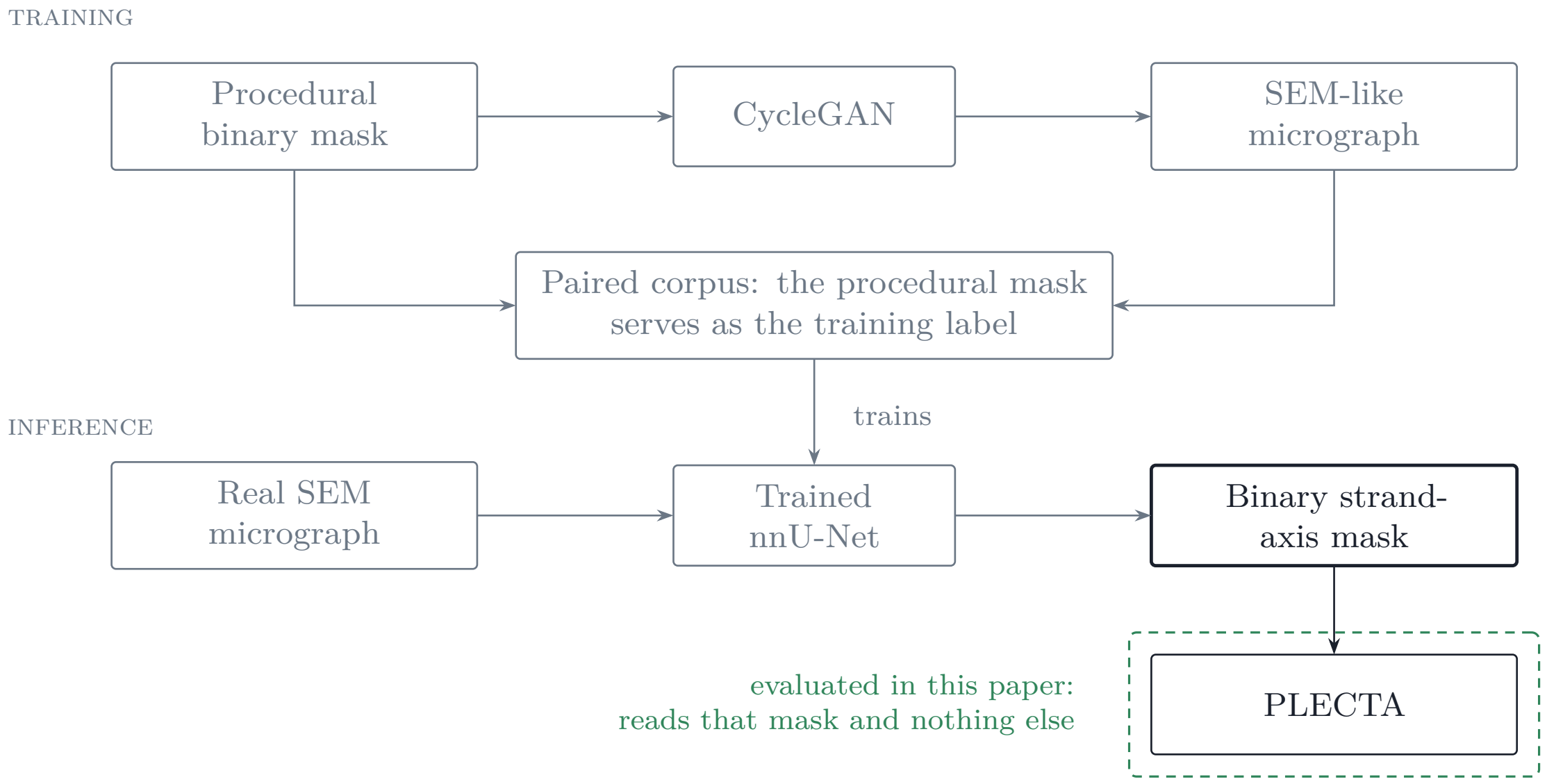
\begin{figure}[!ht]
    \centering
    \input{figures/fig_pipeline_upstream}
    \caption{Upstream generation of the binary strand-axis mask. The segmenter
    is trained on procedurally generated image--mask pairs and requires no
    manual annotations for training or inference. Only the PLECTA stages inside
    the dashed boundary are evaluated here; the segmenter has its own
    evaluation \citep{ruehle2021workflow}.}
    \label{fig:pipeline-upstream}
\end{figure}

An upstream segmenter produces the binary strand-axis mask from each SEM
micrograph (Figure~\ref{fig:pipeline-upstream}), and PLECTA uses that mask
alone.

\begin{figure}[tbp]
    \centering
    \includegraphics[width=\linewidth]{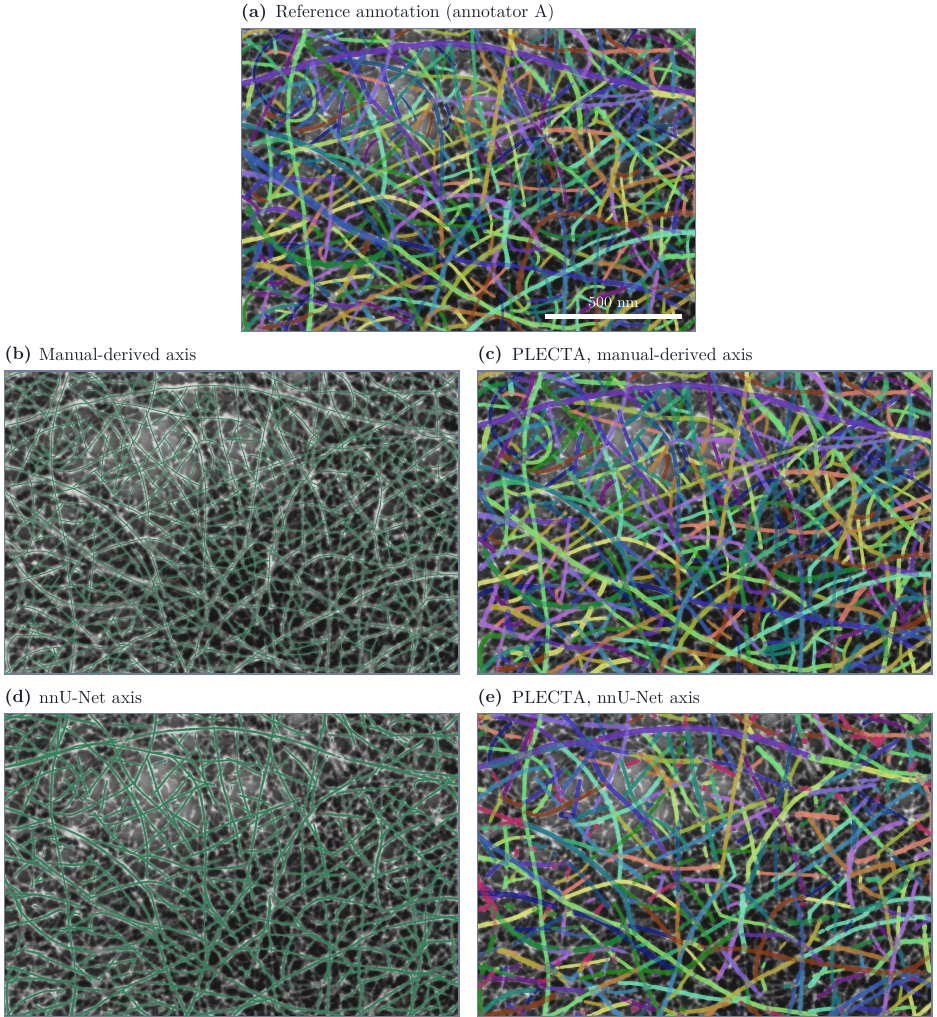}
    \caption{Segmentation of the whole B58\_110 field from two input-mask
    sources; the scale bar in (a) is 500\,nm. B58\_110 is the densest of the
    three fields and the lowest-scoring on the reference-derived axis. \textbf{(a)}~Reference
    annotation; \textbf{(b)}~reference-derived axis; \textbf{(c)}~PLECTA from
    that axis; \textbf{(d)}~automatic axis; and \textbf{(e)}~PLECTA from that
    axis with the same parameters. Input axes are dilated by one pixel for
    legibility. Predicted strands take the colour of their matched reference
    strand; crimson indicates no match.}
    \label{fig:real-sem}
\end{figure}

We evaluated \PlectaRealFieldCountLower{} manually annotated CNT-sheet fields
under two input conditions: the skeleton of the reference annotation and the
automatic nnU-Net axis \citep{ruehle2021workflow}. Both were scored against
the reference annotation, so the conditions differ only in the input mask. The
reference-derived axis tests grouping from curated real-image geometry; the
automatic axis evaluates the complete image-to-instances workflow.
Figure~\ref{fig:real-sem} shows both conditions on one whole field.

\begin{table}[tbp]
    \centering
    \caption{Segmentation from reference-derived and automatic axes in three
    CNT SEM fields. Coverage is the fraction of the frame covered by full-width
    reference strands. Metrics are defined in Section~\ref{sec:mm-metrics};
    detection \Fone{} uses IoU \PlectaDetectionIoU{} and tolerance
    \PlectaPlacementTolerance{}\,px.}
    \label{tab:plecta-real-masks}
    \footnotesize
    \input{results/plecta_real_masks_table}
\end{table}

Table~\ref{tab:plecta-real-masks} reports each field and condition. Mean
\Fone{} was \PlectaRealManualFOneMean{} from the reference-derived axis and
\PlectaRealUNetFOneMean{} from the automatic axis. The reference-derived
result establishes grouping on axes from real micrographs; the automatic
result measures sensitivity to the input mask. With only
\PlectaRealFieldCount{} fields, these results are descriptive and the field is
the independent sample.

The automatic axes were thicker, displaced and more fragmented than the
reference-derived axes (Figure~\ref{fig:real-sem}), and the additional errors
were mainly missed links rather than erroneous mergers. Join \Fone{} narrows
but does not remove the difference between the conditions. On B58\_110, for
example, the automatic axis gave join precision
\PlectaJoinBFiveEightOneOneZeroUNetPrecision{} and recall
\PlectaJoinBFiveEightOneOneZeroUNetRecall{}.

These three fields cannot disentangle segmentation error from ambiguity in
manual strand identity. When PLECTA was given the independent annotation's
skeleton instead of the automatic axis, it achieved
\Fone{}=\PlectaInterAnnotatorBOnAFOneMean{} against the reference. This lies
within the direct inter-annotator agreement range of
\PlectaInterAnnotatorAgreeFLow{}--\PlectaInterAnnotatorAgreeFHigh{}, showing
that plausible manual readings can differ substantially where local occlusions
obscure strand continuity. The automatic-axis result should therefore be
interpreted relative to this reference ambiguity, as discussed in
Section~\ref{sec:limitations}.

\subsection{Component ablations}
\label{sec:res-ablation}

The real-SEM results show the effect of input-mask quality. To isolate the
contribution of PLECTA's segmentation components, we removed one component
at a time on the development scenes. Table~\ref{tab:plecta-ablation} also
includes two controls that replace, rather than remove, part of the method.

The largest losses followed removal of the junction chord-turn term, gap
bridging and junction-cluster consolidation. The remaining rows test the
curvature term, graph construction and iterative refinement; their smaller
effects indicate that these components refine the core segmentation rather
than replace it.

The first control replaces joint junction matching with greedy arm pairing
while retaining the rest of PLECTA, showing the cost of decisions that are
locally cheap but jointly incompatible. The second replaces the linker with a
fixed-gate greedy endpoint-pairing baseline.

\begin{table}[tbp]
    \centering
    \caption{Component ablations and replacement controls on development
    scenes. Each $\Delta F_1$ is relative to PLECTA evaluated on the
    corresponding scene set.}
    \label{tab:plecta-ablation}
    \small
    \input{results/plecta_ablation_table}
\end{table}

\subsection{Comparison with published methods}
\label{sec:res-comparators}

The ablations identify the components that support segmentation. We next
compare the complete method with four published approaches on the
\PlectaGreedySceneCount{} synthetic scenes. The two methods that
accept binary axes were also evaluated on the \PlectaRealFieldCountLower{}
annotated CNT SEM fields of Section~\ref{sec:res-real}. All methods receive
the same binary masks and are scored by the same pipeline. The comparison
therefore concerns overlap-aware grouping of dense, crossing strand axes, not
performance in each method's native application. DNAi \citep{playout2026dnai}, GraFT
\citep{osterlund2026graft} and SIFNE \citep{zhang2017sifne} use released
implementations; \citet{basu2015extraction} is reimplemented because no
implementation was available.

\begin{table}[!ht]
    \centering
    \caption{PLECTA and four published methods on identical masks, ordered by
    \Fone{}. The upper block uses degraded synthetic axes; the lower uses
    reference-derived axes from the CNT SEM fields of
    Section~\ref{sec:res-real}. Variation of information is in bits and lower
    is better; higher is better for the remaining metrics.\\[2pt]
    $^{*}$\,our reimplementation; no implementation of that paper was found.
    \suppcomparators{} details it.}
    \label{tab:plecta-comparators}
    \small
    \input{results/plecta_comparator_table}
\end{table}

For each released method, the mask was supplied at the start of its
identity-assignment stage so that segmentation did not affect the comparison.
All methods therefore received the same masks and were scored by the same
code. Configurations were selected on the same development scenes used for
PLECTA, separate from the scored ones (\suppsplit{}), rather than left at
their defaults. The
ranking also holds on GraFT's independently published generator, reported
below. The injection points and configurations are documented in
\suppcomparators{}.

\begin{figure}[!t]
  \centering
  \includegraphics[width=\linewidth]{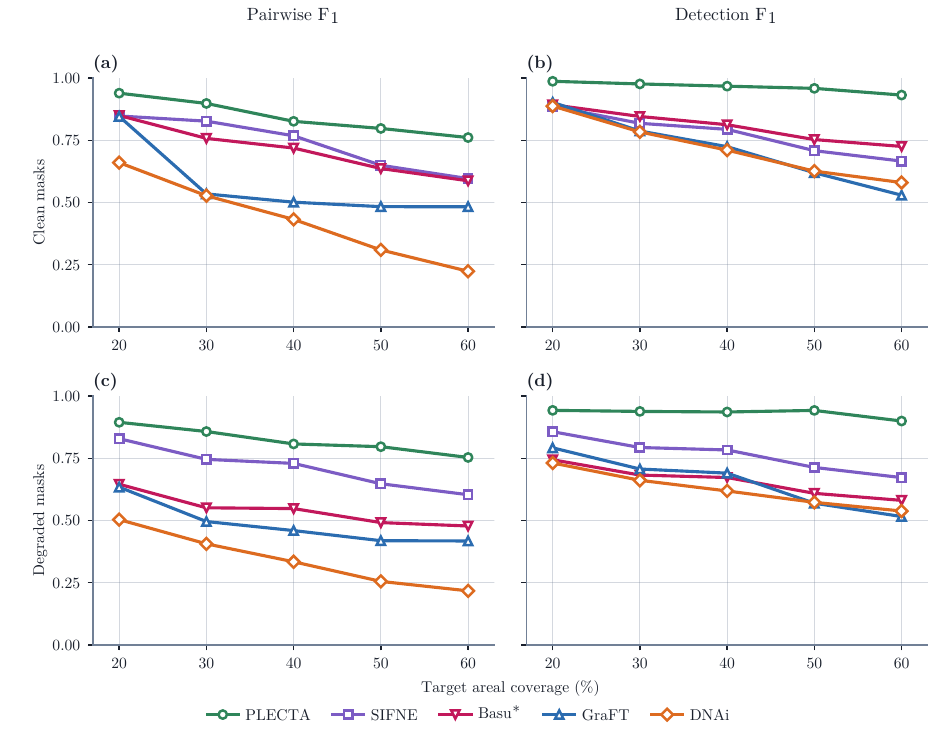}
  \caption{Five methods on identical masks, by coverage. The upper row uses
    clean axes and the lower row degraded axes. Columns: pairwise \Fone{},
    which counts pairs of fragments, and detection \Fone{} at IoU
    \PlectaDetectionIoU{} with a \PlectaPlacementTolerance{}\,px tolerance,
    which counts objects (Section~\ref{sec:mm-metrics}). The two disagree
    where a method fragments, since only the first is quadratic in
    fragmentation. Crossing Fidelity is reported in
    Table~\ref{tab:plecta-comparators}. $^{*}$\,our reimplementation; no
    implementation was available.}
  \label{fig:metric-panels}
\end{figure}

Under these comparator-favouring settings, PLECTA ranked first at every
density and under both mask conditions (Figure~\ref{fig:metric-panels});
SIFNE was the strongest comparator. DNAi was designed for sparse DNA-fibre
micrographs and recovered \PlectaDNAiCleanSparseFOne{} from clean axes at
20\,\% coverage. Basu ranked second on Crossing Fidelity but substantially
lower on pairwise \Fone{}, indicating that curvature-based junction matching
alone is insufficient for the full segmentation task. Detailed results are
given in \suppcomparators{}.

\textbf{Other regimes.} On GraFT's released generator, run at its published
settings over \PlectaRegimeSceneCount{} scenes, PLECTA again achieved the
highest \Fone{} at every density
(\PlectaRegimeOverallPlectaFOne{} overall, compared with
\PlectaRegimeOverallGraFTFOne{} for GraFT and
\PlectaRegimeOverallDNAiFOne{} for DNAi). SIFNE released no generator; on a
recreation of its sparse cobweb topology, PLECTA and SIFNE were similar
at low density and separated only slightly as density increased. This supports
crossing entanglement, rather than strand content alone, as the relevant
difference, although the recreated topology is weaker evidence than a
released generator.

\textbf{Cost.} Table~\ref{tab:plecta-comparators} reports warm,
single-threaded per-scene times. Because the implementations are independently
optimised, these times reflect implementation as well as algorithm.

\textbf{Real SEM.} On reference-derived axes from the
\PlectaRealFieldCountLower{} CNT fields, PLECTA achieved
\Fone{}=\PlectaRealManualFOneMean{}, compared with
\PlectaRealSifneFOneMean{} for SIFNE and \PlectaRealBasuFOneMean{} for Basu,
and ranked highest on every reported measure. The smaller margin than on
synthetic scenes is consistent with these curated axes containing fewer gaps,
fragmentation and placement errors. Together, the ablations and comparisons
show PLECTA's advantage for matched-mask segmentation in dense crossing
networks; the scope of that claim is considered in the conclusions.

%% file: results/plecta_heldout_table.tex
\begin{tabular}{lrrrrrrrr}
\toprule
Group & $n$ & $F_1$ & Precision & Recall & ARI & VI split & VI merge & VI total \\
\midrule
All & 128 & 0.85 & 0.87 & 0.84 & 0.85 & 0.26 & 0.22 & 0.47 \\
20\% & 32 & 0.91 & 0.93 & 0.90 & 0.91 & 0.14 & 0.09 & 0.23 \\
30\% & 32 & 0.87 & 0.89 & 0.86 & 0.87 & 0.21 & 0.16 & 0.37 \\
40\% & 32 & 0.86 & 0.86 & 0.85 & 0.85 & 0.24 & 0.22 & 0.47 \\
60\% & 32 & 0.77 & 0.79 & 0.76 & 0.77 & 0.44 & 0.39 & 0.82 \\
\bottomrule
\end{tabular}

%% file: results/plecta_depth_order_table.tex
\begin{tabular}{lcccc}
\toprule
& \multicolumn{2}{c}{Given 2-D geometry} & \multicolumn{2}{c}{Segmented 2-D} \\
\cmidrule(lr){2-3}\cmidrule(lr){4-5}
Areal coverage & 20\,\% & 40\,\% & 20\,\% & 40\,\% \\
\midrule
\multicolumn{5}{@{}l}{\emph{At each crossing}} \\
Projected crossings per scene & 28 & 150 & 28 & 150 \\
Crossings recovered & 1.00 & 1.00 & 0.86 & 0.75 \\
Order accuracy, all crossings & 0.81 & 0.74 & 0.73 & 0.56 \\
\quad restricted to crossings the stage decided & 0.89 & 0.88 & 0.96 & 0.91 \\
Declined, of crossings the stage found & 0.08 & 0.16 & 0.12 & 0.18 \\
\addlinespace
\multicolumn{5}{@{}l}{\emph{In the resulting stack}} \\
Depth order, all strand pairs & 0.67 & 0.71 & 0.70 & 0.72 \\
\quad restricted to pairs that cross & 0.86 & 0.86 & 0.91 & 0.85 \\
Pairs within one crossing component & 0.39 & 0.84 & 0.36 & 0.75 \\
Layers in the reference & 5.2 & 8.7 & 5.2 & 8.7 \\
Layers recovered & 5.2 & 8.3 & 5.3 & 8.0 \\
Layer agreement & 0.66 & 0.42 & 0.70 & 0.38 \\
\bottomrule
\end{tabular}

%% file: figures/fig_pipeline_upstream.tex
\begin{tikzpicture}[pk, x=1mm, y=1mm]

  \node[ext=32mm] (pmask)   at ( 30,   0) {Procedural\\binary mask};
  \node[ext=24mm] (gan)     at ( 79.5, 0) {CycleGAN};
  \node[ext=32mm] (semlike) at (129,   0) {SEM-like\\micrograph};

  \draw[flowsoft] (pmask) -- (gan);
  \draw[flowsoft] (gan)   -- (semlike);

  \node[ext=54mm] (corpus) at (79.5, -18)
    {Paired corpus: the procedural mask\\serves as the training label};

  \draw[flowsoft] (pmask.south)   |- (corpus.west);
  \draw[flowsoft] (semlike.south) |- (corpus.east);

  \node[ext=32mm] (sem)    at ( 30,  -38) {Real SEM\\micrograph};
  \node[ext=24mm] (nnunet) at ( 79.5,-38) {Trained\\nnU-Net};
  \node[io=32mm]  (bmask)  at (129,  -38) {Binary strand-\\axis mask};

  \draw[flowsoft] (sem)    -- (nnunet);
  \draw[flowsoft] (nnunet) -- (bmask);
  \draw[flowsoft] (corpus.south) -- (nnunet.north);
  \node[note, anchor=west] at (82.5, -28.5) {trains};

  \node[stage=32mm] (plecta) at (129, -56) {PLECTA};
  \draw[flow] (bmask) -- (plecta);

  \begin{scope}[on background layer]
    \node[scope, fit=(plecta)] (core) {};
  \end{scope}
  \node[scopelab, anchor=east, align=right] at (105, -56)
    {evaluated in this paper:\\reads that mask and nothing else};

  \node[note, anchor=south west] at (2,   8) {\textsc{training}};
  \node[note, anchor=south west] at (2, -31) {\textsc{inference}};

\end{tikzpicture}

%% file: results/plecta_real_masks_table.tex
\begin{tabular*}{\linewidth}{@{\extracolsep{\fill}}lrlrrr@{}}
\toprule
Field & Coverage & Input axis & $F_1$ & Join $F_1$ & Det.\ $F_1$ \\
\midrule
B58\_100 & 44.0\,\% & Manual-derived & 0.89 & 0.93 & 0.94 \\
 &  & U-Net & 0.46 & 0.56 & 0.63 \\
\addlinespace
B58\_110 & 46.7\,\% & Manual-derived & 0.84 & 0.91 & 0.93 \\
 &  & U-Net & 0.45 & 0.59 & 0.60 \\
\addlinespace
B58\_300 & 41.5\,\% & Manual-derived & 0.95 & 0.96 & 0.96 \\
 &  & U-Net & 0.47 & 0.62 & 0.58 \\
\bottomrule
\end{tabular*}

%% file: results/plecta_ablation_table.tex
\begin{tabular*}{\linewidth}{@{\extracolsep{\fill}}lrrr@{}}
\toprule
Variant & $F_1$ & $\Delta F_1$ & ARI \\
\midrule
Fixed configuration & 0.87 & +0.000 & 0.86 \\
No joint junction solve & 0.78 & -0.080 & 0.77 \\
No gap bridging & 0.73 & -0.139 & 0.72 \\
Single round & 0.85 & -0.020 & 0.84 \\
No chain-extended frames & 0.85 & -0.021 & 0.84 \\
No round schedule & 0.86 & -0.004 & 0.86 \\
No curvature term & 0.86 & -0.005 & 0.86 \\
No short-connector node merging & 0.85 & -0.022 & 0.84 \\
No junction-cluster consolidation & 0.80 & -0.063 & 0.80 \\
No junction chord-turn term & 0.68 & -0.186 & 0.67 \\
Whole linker replaced & 0.68 & -0.145 & 0.67 \\
\bottomrule
\end{tabular*}

%% file: results/plecta_comparator_table.tex
\begin{tabular*}{\linewidth}{@{\extracolsep{\fill}}lrrrrrrrr@{}}
\toprule
& & & & & & Crossing & \multicolumn{2}{c}{Median s/scene} \\
\cmidrule(l){8-9}
Method & $F_1$ & Precision & Recall & ARI & VI total~$\downarrow$ & fidelity & at 20\,\% & at 60\,\% \\
\midrule
\multicolumn{9}{@{}l}{\emph{Synthetic scenes, degraded axis}} \\
PLECTA & 0.82 & 0.83 & 0.81 & 0.82 & 0.57 & 0.81 & 0.14 & 4.09 \\
SIFNE & 0.71 & 0.77 & 0.67 & 0.70 & 0.93 & 0.77 & 3.42 & 5.68 \\
Basu$^{*}$ & 0.54 & 0.66 & 0.47 & 0.53 & 1.37 & 0.76 & 2.56 & 54.20 \\
GraFT & 0.49 & 0.67 & 0.40 & 0.48 & 1.56 & 0.68 & 2.44 & 29.74 \\
DNAi & 0.34 & 0.46 & 0.28 & 0.33 & 2.11 & 0.58 & 0.04 & 0.56 \\
\addlinespace
\multicolumn{9}{@{}l}{\emph{Real CNT SEM fields, manual-derived axis}} \\
PLECTA & 0.89 & 0.86 & 0.93 & 0.89 & 0.33 & 0.93 &  &  \\
SIFNE & 0.87 & 0.86 & 0.88 & 0.86 & 0.37 & 0.92 &  &  \\
Basu$^{*}$ & 0.76 & 0.74 & 0.78 & 0.76 & 0.59 & 0.89 &  &  \\
\bottomrule
\end{tabular*}

%% file: sections/04_discussion.tex
\section{Conclusions}
\label{sec:conclusions}
\label{sec:discussion}
\label{sec:limitations}

Micrographs of strand networks show crossings that merge distinct strands and
gaps that split one strand into fragments, so each strand's identity is lost
in the image. Recovering it is what makes per-strand length, orientation and
curvature measurable. PLECTA recovers it from a binary image marking strand
centrelines alone, deciding at each junction which segments continue into one
another and allowing a strand to end rather than be forced onward. Crossing
pixels are assigned to every strand passing through them. PLECTA's grouping
core requires no training data and produces identical output from identical
input masks. The upstream segmenter is trained on synthetic image--mask pairs,
so the full workflow requires no manually annotated training micrographs.

\begin{itemize}[leftmargin=1.5em,label=\textbullet,topsep=6pt,itemsep=2pt,parsep=0pt]
  \item Across \PlectaHeldoutSceneCount{} synthetic scenes, PLECTA achieved a mean
        pairwise \Fone{} (agreement on whether two fragments belong to the
        same strand) of \PlectaHeldoutFOneMean{} [\PlectaHeldoutFOneCILow{},
        \PlectaHeldoutFOneCIHigh{}], falling from
        \PlectaDensityTwoZeroFOneMean{} to \PlectaDensitySixZeroFOneMean{} as
        areal coverage, the fraction of the frame the strands fill, rises from 20
        to 60\,\%.
  \item With every method given the same input mask, PLECTA achieved the highest
        scores among the evaluated methods across the tested densities. Its lead
        was also retained on GraFT's released generator at published settings.
  \item The ablation analysis identified the junction term penalising turning, and
        the reconnection of fragments across signal gaps, as the largest
        contributors to segmentation performance.
  \item The greyscale depth stage ordered the crossings it did not decline with
        \PlectaDepthDecidedLow{}--\PlectaDepthDecidedHigh{} accuracy, against
        chance level \PlectaDepthChance{}. Its global order is limited to
        strands connected through crossings.
\end{itemize}

\noindent\textbf{Scope and limitations.}

\begin{itemize}[leftmargin=1.5em,label=\textbullet,topsep=6pt,itemsep=2pt,parsep=0pt]
  \item PLECTA assumes directional continuity through crossings and remains
        sensitive to upstream mask errors. It can bridge short gaps but cannot
        recover omitted strands or reliably correct false axes.
  \item The real-material evaluation is a case study of
        \PlectaRealFieldCountLower{} CNT SEM fields. The automatic-axis mean
        \Fone{} of \PlectaRealUNetFOneMean{} lies within the range over which
        two human annotators agree with each other,
        \PlectaInterAnnotatorAgreeFLow{}--\PlectaInterAnnotatorAgreeFHigh{},
        so it should be interpreted alongside ambiguity in defining strand
        identity through local occlusions.
  \item The comparisons test all methods on the same binary masks in the
        dense, crossing networks studied here. They show PLECTA's advantage in
        this setting, but do not assess each method in the imaging application
        for which it was originally developed.
\end{itemize}

\noindent\textbf{Outlook.}

\begin{itemize}[leftmargin=1.5em,label=\textbullet,topsep=6pt,itemsep=2pt,parsep=0pt]
  \item A more accurate upstream segmenter, trained on synthetic data better
        matched to real micrographs, would improve the masks PLECTA receives.
  \item CNT sheets stacked in a known order, or focused-ion-beam cross-sections
        of the imaged region, would give the true crossing order for the depth
        stage; a real development set, kept separate from the evaluation fields,
        would let parameters be tuned on imaging rather than on synthetic scenes
        alone.
  \item The workflow remains to be tested on other materials, magnifications and
        SEM applications.
  \item The segmented strands, with their depth order and widths, can serve as
        input geometry for mesoscale mechanical simulations.
\end{itemize}